\documentclass[submission,copyright,creativecommons]{eptcs}
\providecommand{\event}{ACL2'14}
\usepackage{breakurl}

\usepackage{amsmath}
\usepackage{amssymb}
\usepackage{booktabs}
\usepackage{graphicx}
\usepackage{listings}
\usepackage{paralist}
\usepackage{subfig}
\usepackage{hyperref}
\usepackage{color}

\title{Initial Experiments with TPTP-style Automated Theorem Provers on ACL2 Problems}
\def\titlerunning{Initial Experiments with TPTP-style Automated Theorem Provers on ACL2 Problems}
\author{
  Sebastiaan Joosten\thanks{Supported by NWO project Effective Layered Verification of Networks on Chips (ELVeN) under grant no. 612.001.108}
  \institute{Technical University of Eindhoven\\Radboud University Nijmegen}
\and
  Cezary Kaliszyk
  \institute{University of Innsbruck}
\and
  Josef Urban\thanks{Supported by NWO grant Knowledge-based Automated Reasoning}\\
  \institute{Radboud University Nijmegen}
}

\def\authorrunning{Joosten et all}

\begin{document}


\maketitle

%
\begin{abstract}
This paper reports our initial experiments with using external ATP
on some corpora built with the ACL2 system.
This is intended to provide the first estimate about
the usefulness of such external reasoning and AI systems for
solving ACL2 problems.
\end{abstract}

\section{Motivation}
\label{introduction}
We are motivated by the recent development of bridges between ITP
systems and their libraries such as Mizar/MML~\cite{mizar-in-a-nutshell}, Isabelle/HOL~\cite{WenzelPN08}, and HOL Light~\cite{Harrison96}/Flyspeck~\cite{Hales05} on one side, 
and
ATP/SMT systems such as E~\cite{Sch02-AICOMM}, Vampire~\cite{Vampire} and Z3~\cite{z3} on the other side. The work
on such systems in the last decade has shown that many top-level
lemmas in the ITP libraries can be re-proved by ATPs after a suitable translation to common ATP formats like TPTP, when given the
right previous lemmas as axioms~\cite{Urb04-MPTP0,MengP08,BlanchetteBPS13,holyhammer}. Automated selection of the right
previous lemmas from such large libraries, and also automated
construction and selection of the right theorem-proving strategies for
proving new conjectures over such large libraries is an interesting AI
problem, where complementary AI methods such as machine learning from
previous proofs and other heuristic selection methods have turned out
to be relatively
successful~\cite{EasyChair:74,HoderV11,KuhlweinLTUH12,abs-1108-3446,MengP09,UrbanV13,US+08,malar14}. Since
2008 the CASC LTB (Large-Theory Batch) competition
division~\cite{Sutcliffe13} has been measuring the performance of
various ATP/AI methods on batches of problems that usually refer to a
large common set of axioms. These problems have been generated
from the large libraries of Mizar, HOL Light and Isabelle, and also
from the libraries of 
common-sense reasoning systems such  as SUMO~\cite{NilesP01} and
Cyc~\cite{RRG05}. This has in the long run motivated the ATP/AI
developers to invent and improve methods and strategies that are useful on
such classes of problems. Such methods then get integrated into strong
advising services (often ``cloud-based'') for the ITP users, such as
HOL(y)Hammer~\cite{hhmcs}, Sledgehammer~\cite{KuhlweinBKU13}, and MizAR~\cite{abs-1109-0616,KaliszykU13b}.

The plan of work that we intend to follow with ACL2 is analogous to
the procedure that was relatively successful with Mizar, Isabelle and
HOL Light: (i) define a translation of the ACL2 formulas to the common
TPTP formats (initially FOF), (ii) export all ACL2 theorems and
definitions into the TPTP format, (iii) find and export the ACL2-proof
dependencies between the ACL2 theorems, (iv) apply machine learning to
such proof dependencies to obtain strong recommendations
(premise-selection) systems that suggest the lemmas useful for proving
a new conjecture, and (v) test the performance of the ATP systems and
their strategies on the problems translated from ACL2, either by
re-proving the theorems from their exact ACL2 proof dependencies, or
by using premise selection to recommend the right lemmas. 
We present the
initial work in these directions and the initial evaluation here.

\section{Translation to TPTP}
Our translation to FOF TPTP is based on the description of the ACL2 logic by Kaufman and Moore~\cite{kaufmann98_acl2}. We add the following axioms defining the ACL2 primitives to each TPTP problem:
\begin{verbatim}
fof(spcax1,axiom, t != nil).
fof(spcax2,axiom, ! [X,Y]: ((X = Y) <=> acleq(X,Y) = t)).
fof(spcax3,axiom, ! [X,Y]: ((X != Y) <=> acleq(X,Y) = nil)).
fof(spcax4,axiom, ! [B,C]: (if(nil,B,C) = C)).
fof(spcax5,axiom, ! [A,B,C]: ((A != nil) => if(A,B,C) = B)).
fof(spcax6,axiom, ! [A]: (not(A) = if(A, nil, t))).
fof(spcax7,axiom, ! [P,Q]: (implies(P,Q) = if(P,if(Q,t,nil),t))).
fof(spcax8,axiom, ! [P,Q]: (iff(P,Q) = and(implies(P,Q),implies(Q,P)))).
fof(and,axiom, ! [A,B]: (and(A,B) = if(A,B,nil))).
fof(or,axiom, ! [A,B]: or(A,B) = if(A,A,B)).
fof(consp1, axiom, ! [A,B]: acleq(consp(cons(A,B)),t) != nil).
fof(consp2, axiom, ! [X]: or(acleq(consp(X),t),acleq(consp(X),nil)) != nil).
fof(consp3, axiom, ! [X]: implies(consp(X),acleq(cons(car(X),cdr(X)),X)) != nil).
\end{verbatim}
and we turn every formula into a TPTP statement by requiring that it
is not equal to nil. 

The translation from ACL2 to TPTP FOF proceeds in two stages.
Initially, for each ACL2 file, after it is loaded and checked by ACL2,
all formulas and proof dependencies are written in the common ACL2
(Lisp-based) format. This is done by a small Lisp (ACL2) program. The
generated Lisp format is next translated to
TPTP by a short Emacs Lisp program.

\subsection{Extraction of formulas and proof dependencies}
One of the strengths of ACL2 is that the user has the ability to use macros.
Many function definitions and theorems are abbreviated through this use of macros.
For getting formulas from ACL2, this poses a problem, since the macros can contain anything, while they are not part of any logic.
To work around this, the books are loaded in ACL2 using the LD command.
After this, we enter raw lisp mode, save a copy of the world, and extract information from it.

We export the \verb=PROOF-SUPPORTERS-ALIST= to be used as input for
the machine learning methods that learn from previous proofs dependencies.
In addition, we export the \verb=THEOREM= and the \verb=LEMMAS= for all symbols in that list, in the order in which they are added to ACL2's logical world (oldest to newest).
As a consequence, macros do not occur in our export anymore.
On the downside, the theorems and definitions that are local to an encapsulate do not occur either.
In addition, no theorem will state that functions terminate.
To make our export to TPTP a bit easier, lambda terms are written out using a function similar to \verb=REMOVE-LAMBDAS=.

An alternative to getting formulas from the ACL2 world, is by using the \verb=SET-OVERRIDE-HINTS=.
Every time the code is called with an empty history, we output the clause to a file.
One drawback is that this way of exporting may be overridden, but we found that this rarely happens in practice.
We plan to use this data as well for learning which hints should be
used for proving a theorem. Such learning could likely be used to
directly advise ACL2 (or its users), but perhaps also to advise the
TPTP-style ATPs running on the translated data.

As far as we are aware, these are the two main options for extracting theorems and lemmas from ACL2 after they are interpreted.
For this paper the data from \verb=PROOF-SUPPORTERS-ALIST= was used.

\subsection{Lisp to Prolog}
The macro-expanded and lambda-expanded formulas in the saved world are
processed by an external Emacs Lisp program that mainly transforms
the Lisp syntax into Prolog (TPTP) syntax. For example the following ACL2 formulas:
\begin{verbatim}
(EQUAL (CDR (CONS X Y)) Y)

(EQUAL (E0-ORDINALP ACL2::X)
       (IF (CONSP ACL2::X)
         (IF (E0-ORDINALP (CAR ACL2::X))
           (IF (EQUAL (CAR ACL2::X) '0)
             'NIL
             (IF (E0-ORDINALP (CDR ACL2::X))
               (IF (CONSP (CDR ACL2::X))
                 (IF (E0-ORD-< (CAR ACL2::X)
                               (CAR (CDR ACL2::X)))
                   'NIL 'T) 'T) 'NIL)) 'NIL)
         (IF (INTEGERP ACL2::X)
           (IF (< ACL2::X '0) 'NIL 'T) 'NIL)))
\end{verbatim}
get converted into these TPTP formulas:
\begin{verbatim}
fof(cdr_cons,axiom,(acleq(cdr(cons(X,Y)),Y) != nil)).

fof(e0_ordinalp,axiom,
   (acleq(e0_ordinalp(X),
          if(consp(X),
             if(e0_ordinalp(car(X)),
                if(acleq(car(X),0),nil,
                   if(e0_ordinalp(cdr(X)),
                      if(consp(cdr(X)),
                         if('e0_ord_<'(car(X),car(cdr(X))),nil,t),t),nil)),nil),
             if(integerp(X),if('<'(X,0),nil,t),nil)))
    != nil)).
\end{verbatim}

We have briefly experimented with more ambitious encodings that
translate ACL2 macros such as AND, OR, IMPLIES, IFF into the
corresponding TPTP connectives and which translate the ACL2 equality
directly as the TPTP equality, however this leads to various initial
problems caused by the fact that everything is an untyped function in
ACL2, while in TPTP one cannot apply a function to a predicate. Still,
such more advanced encodings might result in better performance in
the future. Other issues of the translation to TPTP that we are
gradually addressing in less or more complete ways include the
treatment of quoting, detection of what is a constant and what is a
variable in ACL2, handling of various built-ins such as numbers and
strings, etc. A major source of incompleteness mentioned below is our
current (non-)treatment of ACL2 encapsulation and functional instances.
A particularly useful recently introduced feature of
TPTP is the possibility to quote non-Prolog atoms in apostrophes. This
allows us to re-use practically all ACL2 names in TPTP without any
more complicated syntactic transformations. This is not possible for
variables, whose TPTP names get autogenerated when their ACL2 syntax is
not usable. We use the tptp4X preprocessing tool to collect and universally quantify
all variables in the translated formulas, and also to change the
symbol names when one symbol appears with different arities (this may
happen due to quoting in ACL2). All the extracted TPTP formulas and
their ACL2 dependencies are available at our web
page.\footnote{\url{http://mws.cs.ru.nl/~urban/acl2/00allfof3} and \url{http://mws.cs.ru.nl/~urban/acl2/00alldeps2}}

\section{Reproving Experiments}\label{sec:reprove}

In order to evaluate the completeness of our translation and the initial
performance of TPTP provers, the ACL2 exporter was run on all ACL2
books (directory books/), which took about 800 minutes. These books
contain 3776 Lisp files, of which 1280 can be successfully
processed and translated to TPTP by the current version of our
exporter and translator, producing at least one TPTP formula.
This results in 25,310 unique TPTP FOF. Some of these formulas
originate from ACL2 definitions, hence we use them only as TPTP
axioms and do not consider them for reproving.

For each proved problem we generate a TPTP file that includes the problem as a conjecture
and includes the statements of all the theorems and definitions used by ACL2 in the
proof of the theorem. This gives rise to 23,559 problems. In an initial experiment on
the subset of these problems we evaluated a number of ATPs and their versions choosing
a set of three complementary ones to run on the whole set. The provers and versions
chosen are: Vampire 2.6, E-Prover 1.8 run with
alternate strategy scheduling (BliStr)~\cite{blistr}, and CVC 4. In
order to evaluate the incompleteness of the encoding we also included
the finite counter-model finder Paradox 4. All experiments are run with 10s
time limit on a 48-core server with AMD Opteron 6174 2.2 GHz CPUs, 320 GB RAM, and
0.5 MB L2 cache per CPU and each problem is assigned one CPU.

\begin{figure}[t!]
  \centering
  \begin{tabular}{ccccc}\toprule
  Prover   & Proved (\%)  & Disproved (\%) & Unique & SotAC \\ \midrule
  Vampire  & 4,502 (19.1) &                      & 55 & 0.36 \\
  CVC      & 4,438 (18.8) &                      & 108 & 0.37 \\
  E-Prover & 4,422 (18.8) & \phantom{1}3,826 (16.2)      & 67 & 0.19 \\
  Paradox  & \phantom{10 (0.0)}  & 13,866 (58.9) & & \phantom{0.00} \\ \midrule
  any      & 4,844 (20.6)        & 13,916 (59.1) &  \\\bottomrule
  \end{tabular}
  \caption{\label{fig:hd}Reproving with 10s (23,559 problems)}
\end{figure}

The results of the first reproving experiment are presented in Fig.~\ref{fig:hd},
the second and third columns correspond to Theorem and CounterSatisfiable problem
status returned by the ATP and the last column shows the state of the art
contribution of each prover.
We attribute the high number of disproved theorems to the
incompleteness of our translation. We have been able to identify three main reasons
for this incompleteness.
First, we do not handle encapsulation properly. This means that
proofs that use functional instances cannot be directly replayed by
ATPs. Second, we do not normalize the arities of the functions and
predicates. Functions that are applied with different arities in ACL2
are translated to different constants in the TPTP encoding. This
means that such proofs cannot be currently replayed. Third, ACL2 uses
arithmetic of the underlying system directly. We currently encode all
numeric constants as different TPTP constants.

All three issues can be handled efficiently. For the first two, we
intend to use a translation to a higher-order representation that is immediately
lambda-lifted.
For the third issue, it is possible to translate either to SMT solvers
or the TFA language and provers such SPASS+T.

The reproving results differ quite significantly, depending on the
book. We present the results for particular books, categorized by ACL2
top-level directories in Fig.~\ref{fig:books}. Only book categories
with more than 200 TPTP problems are displayed. Contrary to our expectations,
the arithmetic books have an average proved rate; while datastructures, defsort,
sorting, or models
have quite high disprove rate.

\begin{figure}[t!]
\begin{center}\begin{tabular}{cccccc}\toprule
  Book category    & Proved (\%) & Disproved (\%) & Size \\ \midrule
regex & 35.98 & 51.87 & 667 \\ 
coi & 25.61 & 51.64 & 4810 \\ 
taspi & 24.97 & 52.81 & 905 \\ 
ordinals & 22.30 & 53.15 & 269 \\ 
rtl & 22.06 & 59.61 & 4658 \\ 
centaur & 19.27 & 60.46 & 2848 \\ 
arithmetic-2 & 18.34 & 56.80 & 338 \\ 
arithmetic-5 & 17.71 & 61.14 & 350 \\ 
powerlists & 15.84 & 55.44 & 404 \\ 
defexec & 15.30 & 42.98 & 1019 \\ 
std & 14.25 & 74.96 & 1929 \\ 
arithmetic-3 & 13.11 & 57.37 & 427 \\ 
proofstyles & 12.87 & 81.43 & 264 \\ 
cgen & 12.74 & 60.23 & 259 \\ 
concurrent-programs & 12.19 & 47.38 & 287 \\ 
defsort & 11.74 & 81.37 & 247 \\ 
misc & 10.75 & 60.54 & 958 \\ 
ihs & 9.19 & 83.08 & 337 \\ 
data-structures & 8.80 & 61.12 & 625 \\ 
textbook & 7.40 & 48.14 & 216 \\ 
models & 6.82 & 67.80 & 205 \\ 
\bottomrule
\end{tabular}\end{center}
\caption{\label{fig:books}Reproving results for different book categories (divided by ACL2 top-level directory)}
\end{figure}

\section{Machine Learning Experiments}
To learn which theorems may be relevant to proving other theorems, we
characterize the formulas by the symbols and terms that occur in them
and use such features for measuring the similarity of the
formulas. Similarly to the formulas translated from Mizar, HOL Light
and Isabelle, we can normalize the variables in the terms in various
ways, resulting in various notions of formula simlarity,
see~\cite{holyhammer} for details.
In our ACL2 initial experiments we have focused on the feature
extraction algorithm that performs a unification of variables in
each subterm. With this algorithm, we have extracted 107,944 distinct
features from the ACL2 books, and average number of features per
formula amounts to 17.81.
These feature
numbers (as well as the numbers of training examples - proofs) are
comparable to the numbers that we get from other large libraries, thus
the various efficient sparse machine learning methods developed
recently, e.g., for real-time learning over the whole Mizar
library~\cite{KaliszykU13b} can be used here without any modification.


For each book category, we separately extract the features, dependencies,
and an order of theorems. We use k-NN (k-nearest neighbor) with the
dependencies of each neighbor weighted by the neighbor's distance,
together with
the TF-IDF feature
weighting to produce predictions for every problem. The way the experiments
are done simulates the way a user or ACL2 is developing the library:
at every point, only the previous theorems and their dependencies and
features are known. Based on the features of the conjecture, the k-NN
produces an ordering of the available premises, and a prefix of this order
(the 100 premises that are most likely to help solve the goal) are included
in the TPTP problems.

\begin{figure}[t!]
\begin{center}\begin{tabular}{ccc}\toprule
 Book & 100-Cover (\%) & 100-Precision \\ \midrule
 arithmetic-2 &  94 & 6.84\\ 
 arithmetic-3 &  94 & 6.37\\ 
 arithmetic-5 &  92 & 6.58\\ 
 bdd &  99 & 4.25\\ 
 centaur &  88 & 8.52\\ 
 cgen &  96 & 8.58\\ 
 coi &  89 & 7.33\\ 
 concurrent-programs &  95 & 8.43\\ 
 data-structures &  96 & 7.01\\ 
 defexec &  92 & 9.27\\ 
 defsort &  96 & 6.47\\ 
 ihs &  91 & 8.90\\ 
 misc &  90 & 7.94\\ 
 models &  96 & 9.79\\ 
 ordinals &  94 & 10.64\\ 
 powerlists &  98 & 7.65\\ 
 proofstyles &  96 & 7.24\\ 
 regex &  92 & 4.59\\ 
 rtl &  86 & 10.05\\ 
 std &  93 & 5.36\\ 
 taspi &  95 & 7.60\\ 
 textbook &  95 & 7.39\\ 
\midrule
 all  & 89 & 7.9\\
\bottomrule
\end{tabular}\end{center}
\caption{\label{fig:recall}Machine learning results for different book categories.
For clarity, only books with more than 100 TPTP problems are displayed.}
\end{figure}

We perform two evaluations of the predictions: first a theoretical machine
learning evaluation, second a practical evaluation using TPTP ATPs. For
the theoretical evaluation we measured two factors. \emph{Cover}, expresses how
much of the whole training set is covered in the first 100 predictions.
\emph{Precision}, expresses how many of the first 100 predictions are
from the training set. In the exported set of ACL2 books the average size
of the training example is 12.6 (average number of dependencies); the
100-cover is 89\% and the 100-precision is 7.9. The numbers here again vary
quite significantly depending on the book category, and we present a short
overview in Fig.~\ref{fig:recall}.

Finally we perform an ATP evaluation. The evaluation is performed with the
same setup as the reproving described in Section~\ref{sec:reprove}. For each
problem the 100 k-NN predictions based only on the previous knowledge has
been written to a TPTP file and the four ATP provers are evaluated on the
file. A summary of the results can be found in Fig.~\ref{fig:knn}. The success
rate being about 75--80\% of the dependency reproving rate is in par with the
machine learning premise selection  results in other systems. With a larger
number of advised premises, the number of problems for which finite models
exist has decreased significantly. A number of proofs found by premise
selection concerns the problems which were not provable with the original ACL2
dependencies: the union of the problems solvable is 6,333
(26.9\%). This likely also means that some of the problems that are
currently counter-satisfiable (due to the issues such as
encapsulation) have an alternative first-order proof that goes via
different lemmas.

\begin{figure}[t!]
  \centering
  \begin{tabular}{ccccc}\toprule
  Prover   & Proved (\%)  & Disproved (\%) & Unique & SotAC \\ \midrule
  Vampire  & 3,010 (12.8) &                      & 548  & 0.55 \\
  CVC      & 2,547 (10.8) &                      & 380  & 0.52 \\
  E-Prover & 1,640 (7.0) & 118 (0.5)             & 210  & 0.45 \\
  Paradox  & \phantom{1 (0.0)}   & 2,020 (8.6)   & & \phantom{0.00} \\ \midrule
  any      & 3,749 (15.9)        & 2,020 (8.6) &  \\\bottomrule
  \end{tabular}
  \caption{\label{fig:knn}Machine learning evaluation, 100 predictions with 10s
   run on the 23,559 problems}
\end{figure}

\section{Conclusion}
We have presented the initial experiments with external TPTP-style
ATPs on the large set of problems extracted from the ACL2 books. While
this work is in an early stage, the experiments show that the ATPs can
solve a nontrivial part of the ACL2 lemmas automatically. We hope that
the numbers will still substantially improve, as we address the
remaining reasons for incompleteness.

We also performed an evaluation of machine learning techniques for
premise selection for ACL2 translated problems. The mechanisms
that have been used primarily with ITPs so far, behave quite well with
the ACL2 problems, even without any adjustment. A further tuning of the
algorithms (e.g. adjusting the number of premises, ATP strategies,
machine learning methods, feature extraction, etc.) will likely increase the
proved theorems rates.

Apart from that, there are a number of possible future steps, such as
implementing good proof reconstruction techniques for importing the
ATP proofs, trying specialized inductive provers (e.g., the QuodLibet
system, etc.), or even employing ACL2 via the TPTP framework as a
strong inductive ATP that helps other ITPs with problems where
induction is needed. A by-product of this work are the large number of
TPTP problems, on which the general ATP and TPTP/CASC community can
try to improve their methods.\footnote{
All the problems and data generated in our evaluation are available at
  \url{http://cl-informatik.uibk.ac.at/users/cek/acl2datasel/}.
The re-proving problems (using acl2-dependencies) are in the file reprove.tgz,
the k-NN prediction problems we evaluated are in the file knn.tgz, 
and the data packed for each book directory separately are in the file books.tgz.}

\bibliography{ate11}
\bibliographystyle{eptcs}

\end{document}